%% file: main.tex
\documentclass[submission,copyright,creativecommons,spaces,hyphens]{eptcs}
\providecommand{\event}{SNR 2021} 
\usepackage{breakurl}             
\usepackage{underscore}           
\usepackage{cite}
\usepackage{amsmath,amssymb,amsfonts}
\usepackage{algorithmic}
\usepackage{graphicx}
\usepackage{textcomp}
\usepackage{xcolor}
\usepackage{breqn}
\usepackage{caption}
\usepackage{subcaption}
\usepackage{todonotes}
\usepackage[font=small]{caption}

\newcommand{\figref}[1]{Fig.~\ref{fig:#1}}
\newcommand{\figreftwo}[2]{Figs.~\ref{fig:#1}~and~\ref{fig:#2}}

\title{Work In Progress: Safety and Robustness Verification of Autoencoder-Based Regression Models using the NNV Tool}
\author{Neelanjana Pal
\institute{Department of Electrical and Computer Engineering\\
Vanderbilt University, USA}
\email{neelanjana.pal@vanderbilt.edu}
\and
Taylor T Johnson \institute{Department of Electrical and Computer Engineering\\
Vanderbilt University, USA}
\email{taylor.johnson@vanderbilt.edu}
}
\def\titlerunning{Safety and Robustness Verification of Autoencoder-Based Regression Models using the NNV Tool}
\def\authorrunning{Neelanjana Pal, Taylor T Johnson}
\begin{document}
\maketitle

\begin{abstract}
\input{abstract}
\end{abstract}

\input{introduction.tex}

\input{RelatedWork.tex}
\input{Background.tex}
\input{projectDefinition}
\input{projectProposal} 
\input{conclusion}

\section*{Acknowledgements}
The material presented in this paper is based upon work supported by the National Science Foundation (NSF) through grant numbers 1910017, 1918450, and 2028001, the Defense Advanced Research Projects Agency (DARPA) under contract number FA8750-18-C-0089, and the Air Force Office of Scientific Research (AFOSR) under contract number FA9550-22-1-0019.
Any opinions, findings, and conclusions or recommendations expressed in this work in progress publication are those of the authors and do not necessarily reflect the views of AFOSR, DARPA, or NSF.

\bibliographystyle{eptcs}
\bibliography{references}

\end{document}

%% file: abstract.tex
This work in progress paper introduces robustness verification for autoencoder-based regression neural network (NN) models, following state-of-the-art approaches for robustness verification of image classification NNs. Despite the ongoing progress in developing verification methods for safety and robustness in various deep neural networks (DNNs), robustness checking of autoencoder models has not yet been considered. We explore this open space of research and check ways to bridge the gap between existing DNN verification methods by extending existing robustness analysis methods for such autoencoder networks. While classification models using autoencoders work more or less similar to image classification NNs, the functionality of regression models is distinctly different. We introduce two definitions of robustness evaluation metrics for autoencoder-based regression models, specifically the percentage robustness and un-robustness grade. We also modified the existing Imagestar approach, adjusting the variables to take care of the specific input types for regression networks. The approach is implemented as an extension of NNV, then applied and evaluated on a dataset, with a case study experiment shown using the same dataset. As per the authors' understanding, this work in progress paper is the first to show possible reachability analysis of autoencoder-based NNs.

%% file: introduction.tex
\section{Introduction}
\label{sec: Introduction}

State-of-the-art and well-trained neural networks (NN) can easily be attacked by small perturbations in inputs, leading to significant aberrations in their outputs \cite{moosavi2016deepfool,LBFGS,goodfellow2014explaining}. These input perturbations are not only limited to image-based networks but also apply to other input types as well, e.g., time-series data or input signals. Such lack of robustness poses serious risks to information integrity, privacy and security, and can be catastrophic in safety-critical applications \cite{saxena2017design,elattar2019conception}. While verification of NNs with image inputs is a vastly growing research area; specifically, with recent ongoing works on safety and robustness checking of feedforward (FFNN), convolutional (CNN), and semantic segmentation networks (SSN); less has been done in the domain of autoencoder verification. Classification models using autoencoders work almost similar to usual classifiers, but there is a need for new research to develop verification techniques for regression models. The regression-based autoencoders regenerate the input in its output and thus can be checked using verification techniques whether the recreated output comes within a certain accepted range of the unperturbed input, in case there is a certain fault/attack on its input side.

In a prior work, the authors of \cite{tran2019fm} introduced a novel  framework for NN verification named \textbf{Neural Network Verification} (NNV)\cite{tran2020nnv} tool, capable of evaluating the robustness of several DNN architectures, e.g., FFNN, CNN, SSN, etc. Later, a new set-based approach,\textbf{ Imagestar} \cite{tran2019fm,tran2020verification} is also incorporated into this tool. In this work in progress work, we explore similar methods in the context of autoencoder verification via experimenting on a sampled dataset and checking if the output lies within a pre-determined safe threshold around the corresponding uninterrupted input values, given a specific type of fault in the input.

%% file: RelatedWork.tex
\section{Related Work}
\label{sec: Relatedwork}

Safety verification and robustness checking has attracted enormous attention in many application areas such as machine learning \cite{lomuscio2017approach, kouvaros2018formal, akintunde2018reachability, singh2018fast, akintunde2019verification, wang2018formal,weng2018towards,zhang2018efficient}, formal methods \cite{pulina2010abstraction,katz2017reluplex, tran2019parallel, xiang2018specification,xiang2017reachable, huang2017safety,dutta2018output,xiang2018output}, and security \cite{gehr2018ai,wang2018formal,wang2018efficient} etc. Many researchers are focused on developing new formal method based tools to address this vastly evolving field, some examples being satisfiability modulo theories (SMT) \cite{katz2017reluplex}, polyhedron \cite{tran2019parallel,xiang2017reachable}, mixed-integer linear program (MILP) \cite{dutta2018output}, interval arithmetic \cite{wang2018formal,wang2018efficient}, zonotope \cite{singh2018fast}, input partition \cite{xiang2018specification}, linearization \cite{weng2018towards}, abstract-domain \cite{singh2019abstract} and star set based \cite{tran2019fm} methods. So far, several safety verification methodologies have been proposed for different neural network (NN) architectures, the focus mostly being on Feed-Forward Networks \cite{tran2019fm,dutta2018learning,xiang2018specification,ehlers2017formal}, Convolutional Neural networks \cite{singh2019abstract,kouvaros2018formal, katz2019marabou, anderson2019optimization,tran2020verification,dvijotham2020efficient}, Semantic segmentation networks \cite{2020Segmentation,full2020studying,Oliveira2017,Klingner,tran2021cav,arnab2018robustness,Zhou2020AutomatedEO} and some on Recurrent Neural Networks \cite{khmelnitsky2020property,akintunde2019verification}. 

Autoencoder NNs are present in the literature for quite some time and their typical application lies in data-denoising \cite{vincent2008extracting,vincent2010stacked}, high quality non-linear feature detection\cite{7954012}, anomaly detection\cite{zhou2017anomaly,sakurada2014anomaly,8363930}, fault classification\cite{sun2016sparse,munir2020performance}, imbalanced data classification\cite{7449810}, etc.; but as per the authors' knowledge there is no prior verification work in the domain of autoencoder-based NNs.

%% file: background.tex
\section{Background}
\label{sec: Background}
\subsection{Autoencoder}
Autoencoders are a subset of Artificial Neural Networks (ANN) that try to reconstruct an input at the output. In general, an autoencoder model is composed of two main components: an encoder and a decoder. The encoder part compresses the input into a latent space representation, and the decoder tries to recreate the input from the encoder output. In other terms, the encoder reduces the feature dimensions of the input (similar to Principal Component Analysis\cite{ringner2008principal}) and can thus be used for the data preparation steps for other machine learning models. Based on the learning objective, these applications can be broadly classified into two main categories:
\begin{enumerate}
    \item Regression Task: Here, the autoencoder is used to recreate the input at the output.
    \item Classification Task: In this application, a complete autoencoder model is first trained as a regression model. Then the decoder part is removed from the model; a softmax and a classification layer are added after the bottleneck, and the input labels are 'one hot-coded' and passed along with the input. This slightly modified model is then trained again to generate an autoencoder-based classification model. 
\end{enumerate}
\begin{figure}[h!]
    \centering
    \includegraphics[width = 0.6\columnwidth]{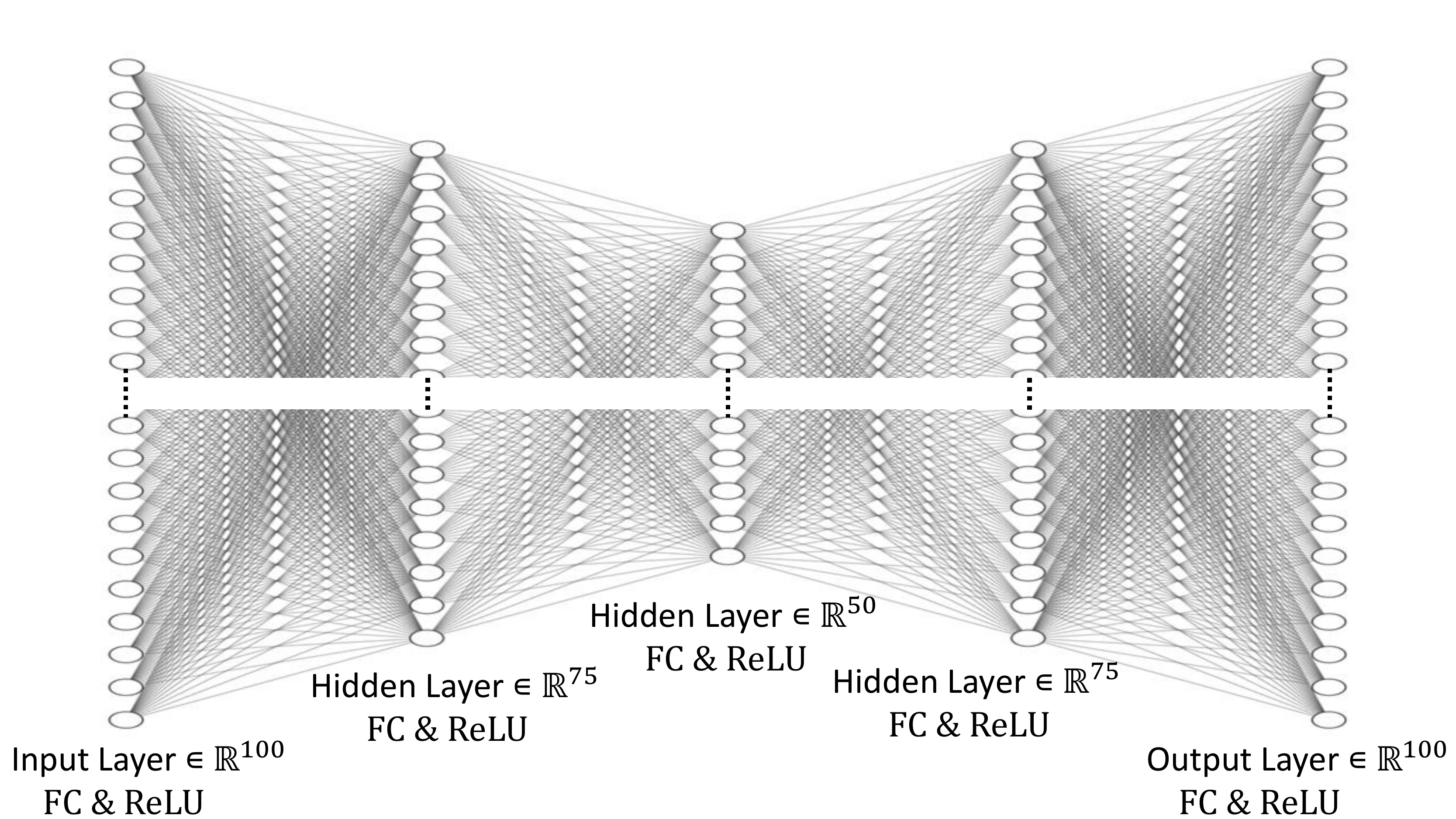}
    \caption{Autoencoder Model used in the paper [*FC: Fully connected]}
    \label{fig:Autoencoder}
\end{figure}

As mentioned in Sec.\ref{sec: Introduction}, classification models using autoencoders works almost similar to a regular classifier, hence the classification task is not included in the scope of this paper.
\subsection{Neural Network Verification Tool\cite{tran2020nnv}}

The Neural Network Verification (NNV) tool is a set-based framework for NN verification~\cite{tran2020nnv}. It supports several reachability algorithms for safety verification and robustness analysis of different types of DNNs. In general, for reachability analysis of a NN, the output reachable sets are obtained in a layer-by-layer manner, from a given input, specified by the upper and lower bounds of perturbation around the actual one. The reachable sets at the final layer are the collection of all possible states of the DNN. The DNN is considered 'safe' iff there is no intersection between the output sets and the unsafe region, defined by the safety properties \cite{tran2020nnv}.

\textbf{Reachability using NNV:} For NN verification with image inputs, a new approach\cite{tran2020verification} was introduced as a part of NNV, called Imagestar. Imagestar is the first set-based approach to efficiently deal with deep CNNs, e.g., VGG16 and VGG19, which combines the operations on images with linear program solving.

The input for the regression model considered in this paper is a set of time-instanced signals. Imagestars~\cite{tran2020verification} are not directly applicable here and require extension. For reachability analysis using NNV, an approach with a similar underlying concept of Imagestar is used, but the dimension of input will change to a 1D equivalent vector, and as the specific input used here will be a signal, we name this approach 'Signalstar'. 

Similar to Imagestar, Signalstar is also a tuple of three variables $\langle c, V, P \rangle $, and the set of signals represented by the Signalstar is given as:
 \begin{equation}
     [\![ \Theta ]\!]= \{x \mid x=c+\sum_{i=1}^{m}{(\alpha_{i}v_{i})} \mid P(\alpha_{1},\alpha_{2},....,\alpha_{m})=\top\}
 \end{equation}
%
where $c \in \mathbb{R}^{n}$ is the anchor signal or central signal with n time instances. Similar to Imagestar $V=\{v_{1},v_{2},...v_{m}\}$ is generator signal instances, which is a set of $m$ signals in $\mathbb{R}^{n}$. The generator signals are arranged to form the basis array of Signalstar $(n\times m)$. $P:\mathbb{R}^{m}\xrightarrow{} \{ \top,\bot \}$ is a predicate.

\begin{figure}[h!]
    \centering
    \includegraphics[width = 0.3\columnwidth]{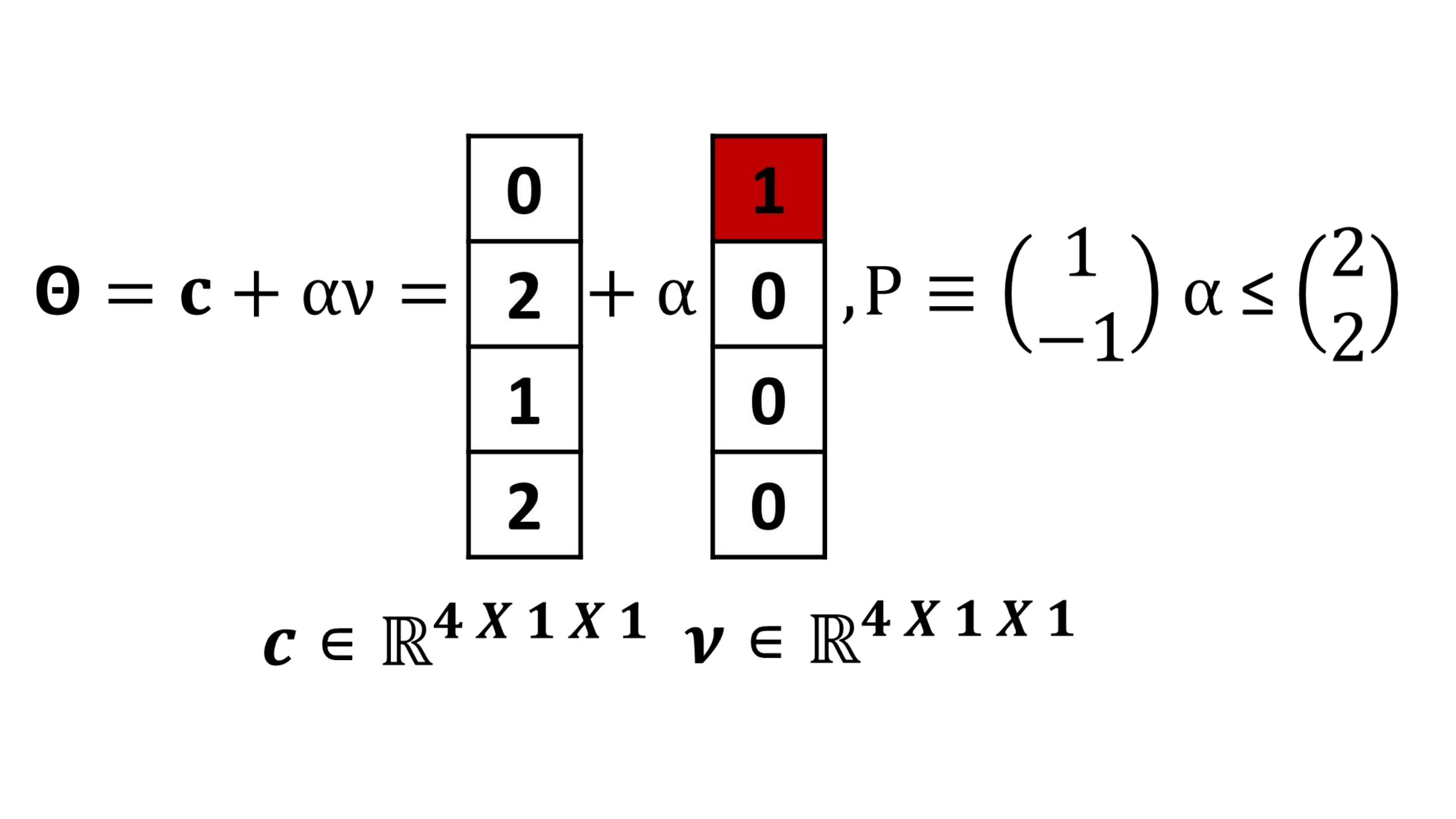}
    \caption{Signalstar}
    \label{fig:Signalstar}
\end{figure}

Another way of defining the Imagestar or Signalstar set is to use the upper and lower bounds of the attack centering the actual input. These bounds on each input parameter along with the predicates create the complete set of constraints the optimizer will solve to generate the initial set of states. An example of Signalstar is shown in Fig. \ref{fig:Signalstar}.

\subsection{Adversarial Attack on Signal or Signal Noise} An adversarial attack can be defined as some unwanted modification in the input, which causes the deviation in output w.r.t the actual result. Here, some signal noises added by the sensors
can be thought of as equivalent to the adversarial attack on the input. For this paper, we will consider a simpler noise, called impulse noise, a random occurrence of spikes of very short duration and relatively high amplitude. These noises are alternatively termed as 'Spike faults'. While coming from a sensor, a common source for this noise is voltage spikes in the device. Because of its sporadic nature, it's really difficult to detect and analyze.

\begin{figure}[h!]
    \centering
    \includegraphics[width = 0.7\columnwidth]{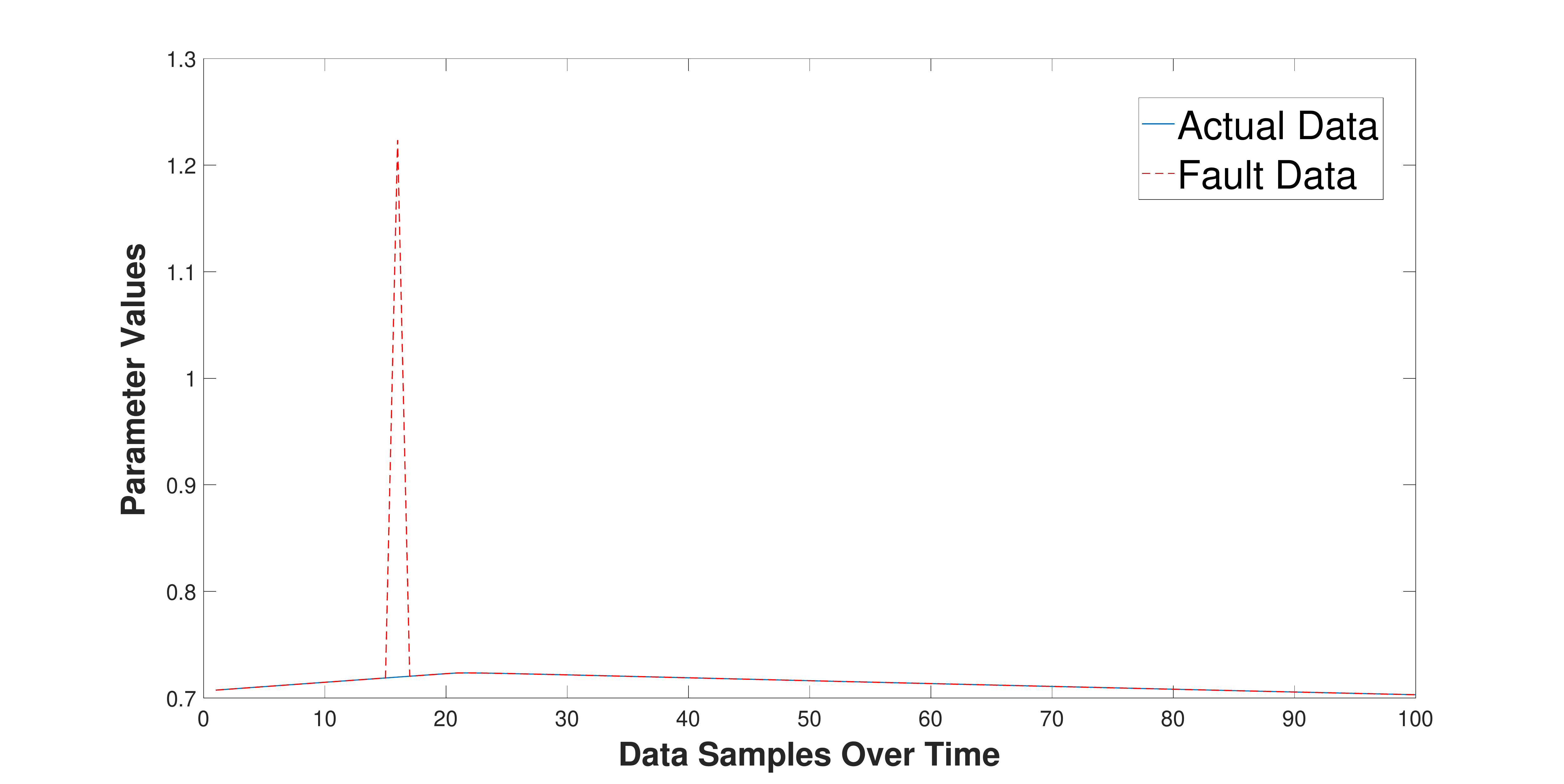}
    \caption{Sample Fault Signal and the actual signal}
    \label{fig:Samplesignal}
\end{figure}

A sample fault signal (in red) overlapping the actual signal (in blue) is shown below in \figref{Samplesignal}

%% file: projectDefinition.tex
\section{Problem Formulation: Verification of Autoencoder Models}
\label{sec:Problem Definition}
 The problem statement is devised with the help of and collaboration with TU Munich\footnote{Hongpeng Cao, Mirco Theile, Bingzhuo Zhong, Dr.Marco Caccamo of \textbf{Chair of Cyber-Physical Systems in Production Engineering, Technical University of Munich (TUM)}}, where, autoencoder verification is an important part of a closed-loop system. It can be described as: first, a sensor collecting a signal with a fixed time range; then, passing the sampled sensor data to the autoencoder to get the best possible version of the time signal. Once the signal is reconstructed, it is sent to a controller for further applications.

\begin{figure}[h!]
    \centering
    \includegraphics[width=\columnwidth]{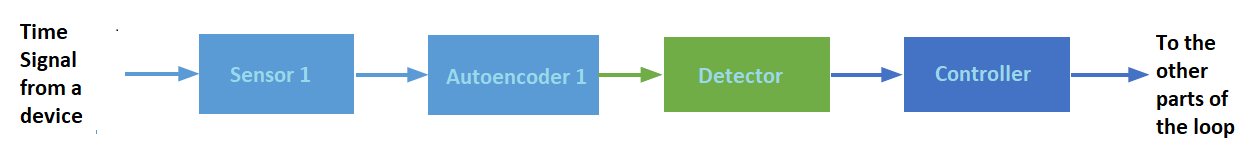}
    \caption{System Model}
    \label{fig:Sysmodel}
\end{figure}

 Here, the task that we are concerned with, is to analyze the reconstructed output of the autoencoder using reachability methods and by defining some measures to protect the controller from faulty inputs. In other words, the evaluation process includes checking how big a fault can be tolerated by the controller, given an uninterrupted signal, the fault location, and an acceptable pair of upper-lower limits. In real life, this range depends on the devices the output signal will be passed to, and their allowable capacity of accepting min and max current/voltage values. The necessary components of the complete loop are as shown in \figref{Sysmodel}.
 
While the data is passed from the sensor to the autoencoder, sensor noises may get added to the original data. As the controller will be designed to work with a specific range of input values, external errors can cause the controller to malfunction and thus can cause the entire closed-loop system to collapse. So it is necessary to detect (in the Detector part of \figref{Sysmodel}) if the reformed signal is within the acceptable range of the controller or not.

%% file: projectProposal.tex
\section{Case Study}
\label{sec:Experimental Setup}

\subsection{\textbf{Experimental Steps}} 
(1) \textbf{Dataset Selection and Autoencoder Model Training:} The dataset  is provided by our collaborators in TU Munich \footnote{Hongpeng Cao, Mirco Theile, Bingzhuo Zhong, Dr. Marco Caccamo of \textbf{Chair of Cyber-Physical Systems in Production Engineering, Technical University of Munich (TUM)}}. Each signal segment consists of 100 time samples and is then normalized using Min-Max technique. The entire dataset is divided into, 2057 training samples and 363 test samples. For generating the fault signals, the same test samples are used with spike faults at random time instances. For the primary phase of work, only one fault per signal is considered as of now.
    
Autoencoder architecture used for this paper is a five-layer NN, shown in \figref{Autoencoder}. The first two layers comprise the encoder part of the architecture and the last two, the decoder part. The third layer creates the latent space for the model. Here, the 'Adam' optimizer and 'Mean square error' loss function is used for training purposes.

\noindent(2) \textbf{Attacks/Perturbations on the Inputs for Verification:} Verification is always done on the inputs with certain noise or perturbations to check if the model can produce the same outputs correctly, i.e., regenerate the unperturbed inputs. As mentioned in Sec.\ref{sec: Background}, spike faults are added for this regression task. 

\noindent(3) \textbf{Reachability Analysis Using NNV:} Given a pair of upper-lower limits of the attack around the central signal, the input set created using Signalstar will propagate through the different layers and sometimes split into multiple sets (because of the presence of the non-linear activation functions/layers). Finally, at output layer produces the collection of output sets. The final output sets thus created are called the reachable sets of the model w.r.t the input attack. Once the output reachable sets are calculated using NNV, it is checked if they intersect with the unsafe regions (specified in terms of the faulty signal). 

\noindent(4) \textbf{Evaluation Measures:} 
In the case of signal inputs, even after the presence of a fault, if the output set can recreate a similar input signal or if the bounds of the output signal are within a certain permissible limit of the input signal, the model is said to be robust. As autoencoder verification is a comparatively new direction of work, the measurement of success is relative. As analysis of safety and robustness verification has already been done on Classification based NNs, we can use a similar measure as well. Though for the regression model the evaluation measure is still an open question, we tried to devise two evaluation metrics for this paper:

(a) \textbf{Percentage Robustness:} For calculating the robustness and safety measure for such application, percentage robustness measure can be used. It can be defined as the number of instances where the output bound is within the threshold values (i.e., the acceptable upper and lower bounds), divided by the total number of time instances. It gives a measure of the relative safety of the network w.r.t the input signal and for a given fault a value 100 being perfectly safe and 0 being completely unsafe.

(b) \textbf{Un-robustness Grade:} This failure grade can be defined as the maximum difference between the actual signal and corresponding output bound (upper or lower) divided by the threshold value (acceptable deviation). The higher it is from value 1, the more it's deviated from the permissible range, i.e., the grade of Un-robustness is more.

\subsection{Experiment Results}

\begin{figure}[h!]
     \centering
     \begin{subfigure}[b]{\columnwidth}
         \centering \includegraphics[width=0.8\columnwidth]{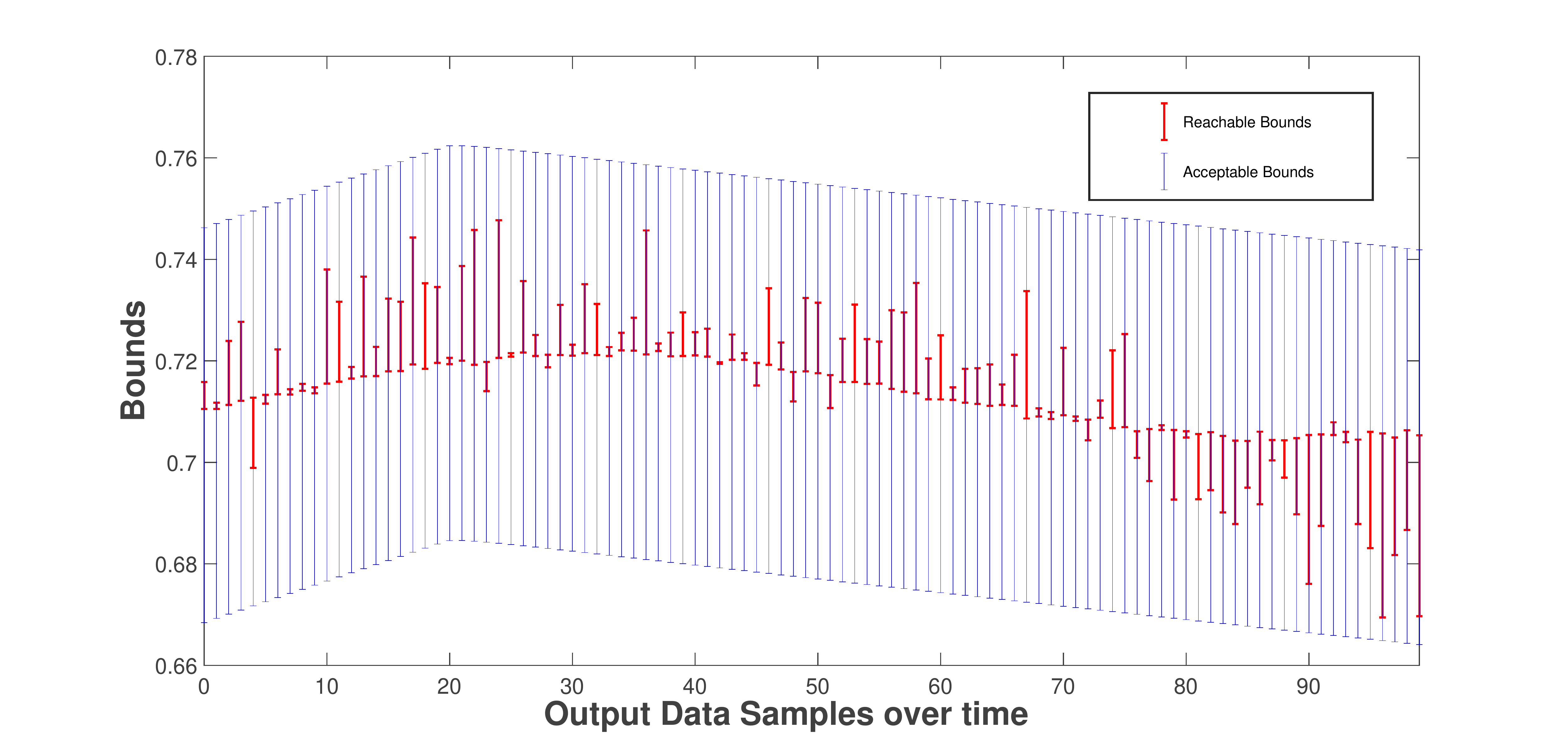}
        \caption{Threshold = 0.0389}
    \label{fig:BoundsThreshold1}
     \end{subfigure}
     \hfill
     \begin{subfigure}[b]{\columnwidth}
         \centering
         \includegraphics[width = 0.8\columnwidth]{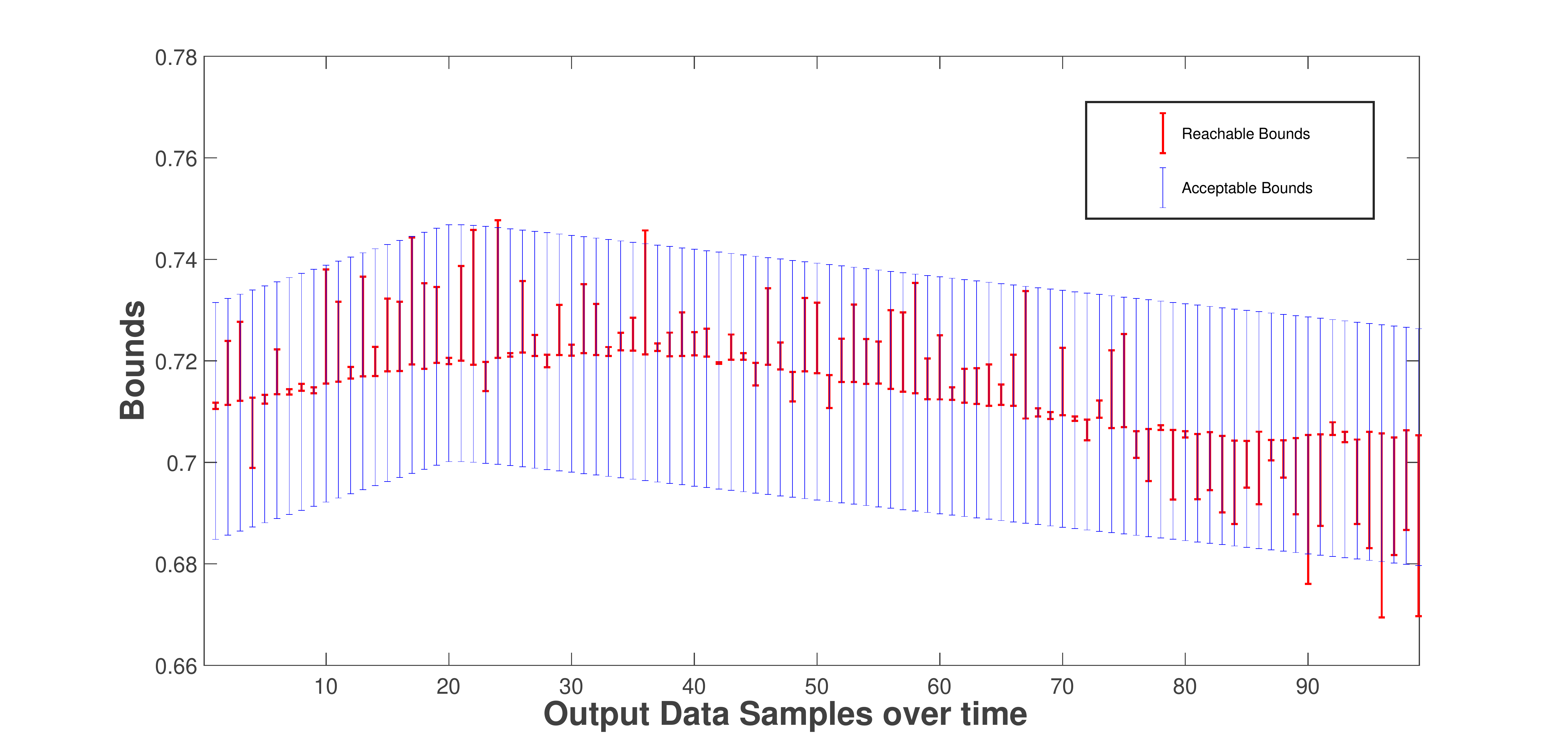}
        \caption{Threshold = 0.0233}
    \label{fig:BoundsThreshold2}
     \end{subfigure}
        \caption{Output bounds (red) of the fault signal and bounds (blue) for the input signal with 2 different thresholds.}
        \label{fig:Bounds2diffThreshold}
\end{figure}
\figreftwo{BoundsThreshold1}{BoundsThreshold2} depict the plots of output bounds corresponding to the sample signals shown in \figref{Samplesignal}, both for the actual and its faulty counterpart; faulty signal bounds at the output being shown in red and the actual signal with permissible limits in blue for two separate acceptable ranges (for example +/-0.0389 and +/-0.0233). We can see from these two plots that whether the reconstructed output signal will be accepted for the controller or not, also depends on this range; while in \figref{BoundsThreshold1}, the same output bounds come well within the range and hence considered robust; bounds for some time instances in \figref{BoundsThreshold2} go beyond the accepted range and hence is not safe to use for further process. For the example, \figref{BoundsThreshold2} previously defined evaluation matrices can be calculated as:

\noindent\textbf{Percentage Robustness:} total time instance = 100; instances, where output bound is within the permissible limit = 95; therefore percentage Robustness = .95.

\noindent\textbf{Un-robustness Grade:} deviation is maximum at time instance 96; here lower bound is deviated maximum from the actual signal and the corresponding value is 0.6695; lower limit of permissible range at t = 96 is 0.7057;hence maximum deviation = 0.7057-0.6695 = 0.0363; finally the Grade of Un-robustness = 0.0363/0.0233 = 1.5536. 


%% file: conclusion.tex
\section{Conclusion}
\label{sec: Conclusion & Future Scope}

In this work in progress paper, we present the first formal verification approach for autoencoders, based on reachability using Imagestars. In our case study, the regression-based autoencoder model takes time-instanced signals from a device, which might have spike faults, and we analyzed the reconstructed output w.r.t an uninterrupted input. Given an accepted value of upper and lower limit (at output), we then checked if the reconstructed signal is within that range of the actual signal. We evaluated the results with two different threshold values. 

Though this is the first stage of analysis of autoencoder based models, in future work, we will continue to develop better evaluation metrics that are more relevant to the use case where reachability-based analysis may be most effective. We can also modify other verification tools used for classification models and investigate comparative effectiveness with them. Incorporating other autoencoder applications as case studies is also planned for our future work. Another avenue can be to study, how the denoising property of an autoencoder help mitigate the effect of input faults in the output, and that in turn increases robustness. Robustness checking of variational autoencoder (VAE) models may also be an interesting direction to explore.

%% file: main.bbl
\begin{thebibliography}{10}
\providecommand{\bibitemdeclare}[2]{}
\providecommand{\surnamestart}{}
\providecommand{\surnameend}{}
\providecommand{\urlprefix}{Available at }
\providecommand{\url}[1]{\texttt{#1}}
\providecommand{\href}[2]{\texttt{#2}}
\providecommand{\urlalt}[2]{\href{#1}{#2}}
\providecommand{\doi}[1]{doi:\urlalt{http://dx.doi.org/#1}{#1}}
\providecommand{\eprint}[1]{arXiv:\urlalt{https://arxiv.org/abs/#1}{#1}}
\providecommand{\bibinfo}[2]{#2}

\bibitemdeclare{inproceedings}{akintunde2018reachability}
\bibitem{akintunde2018reachability}
\bibinfo{author}{Michael \surnamestart Akintunde\surnameend},
  \bibinfo{author}{Alessio \surnamestart Lomuscio\surnameend},
  \bibinfo{author}{Lalit \surnamestart Maganti\surnameend} \&
  \bibinfo{author}{Edoardo \surnamestart Pirovano\surnameend}
  (\bibinfo{year}{2018}): \emph{\bibinfo{title}{Reachability analysis for
  neural agent-environment systems}}.
\newblock In: {\sl \bibinfo{booktitle}{Sixteenth International Conference on
  Principles of Knowledge Representation and Reasoning}}.

\bibitemdeclare{inproceedings}{akintunde2019verification}
\bibitem{akintunde2019verification}
\bibinfo{author}{Michael~E \surnamestart Akintunde\surnameend},
  \bibinfo{author}{Andreea \surnamestart Kevorchian\surnameend},
  \bibinfo{author}{Alessio \surnamestart Lomuscio\surnameend} \&
  \bibinfo{author}{Edoardo \surnamestart Pirovano\surnameend}
  (\bibinfo{year}{2019}): \emph{\bibinfo{title}{Verification of RNN-Based
  Neural Agent-Environment Systems}}.
\newblock In: {\sl \bibinfo{booktitle}{Proceedings of the 33th AAAI Conference
  on Artificial Intelligence (AAAI19). Honolulu, HI, USA. AAAI Press}},
  \doi{10.1609/aaai.v33i01.33016006}.

\bibitemdeclare{inproceedings}{anderson2019optimization}
\bibitem{anderson2019optimization}
\bibinfo{author}{Greg \surnamestart Anderson\surnameend},
  \bibinfo{author}{Shankara \surnamestart Pailoor\surnameend},
  \bibinfo{author}{Isil \surnamestart Dillig\surnameend} \&
  \bibinfo{author}{Swarat \surnamestart Chaudhuri\surnameend}
  (\bibinfo{year}{2019}): \emph{\bibinfo{title}{Optimization and abstraction: a
  synergistic approach for analyzing neural network robustness}}.
\newblock In: {\sl \bibinfo{booktitle}{Proceedings of the 40th ACM SIGPLAN
  Conference on Programming Language Design and Implementation}}, pp.
  \bibinfo{pages}{731--744}, \doi{10.1145/3314221.3314614}.

\bibitemdeclare{inproceedings}{arnab2018robustness}
\bibitem{arnab2018robustness}
\bibinfo{author}{Anurag \surnamestart Arnab\surnameend},
  \bibinfo{author}{Ondrej \surnamestart Miksik\surnameend} \&
  \bibinfo{author}{Philip~HS \surnamestart Torr\surnameend}
  (\bibinfo{year}{2018}): \emph{\bibinfo{title}{On the robustness of semantic
  segmentation models to adversarial attacks}}.
\newblock In: {\sl \bibinfo{booktitle}{Proceedings of the IEEE Conference on
  Computer Vision and Pattern Recognition}}, pp. \bibinfo{pages}{888--897},
  \doi{10.1109/CVPR.2018.00099}.

\bibitemdeclare{article}{7954012}
\bibitem{7954012}
\bibinfo{author}{M.~\surnamestart {Chen}\surnameend},
  \bibinfo{author}{X.~\surnamestart {Shi}\surnameend},
  \bibinfo{author}{Y.~\surnamestart {Zhang}\surnameend},
  \bibinfo{author}{D.~\surnamestart {Wu}\surnameend} \&
  \bibinfo{author}{M.~\surnamestart {Guizani}\surnameend}
  (\bibinfo{year}{2017}): \emph{\bibinfo{title}{Deep Features Learning for
  Medical Image Analysis with Convolutional Autoencoder Neural Network}}.
\newblock {\sl \bibinfo{journal}{IEEE Transactions on Big Data}}, pp.
  \bibinfo{pages}{1--1}, \doi{10.1109/TBDATA.2017.2717439}.

\bibitemdeclare{inproceedings}{8363930}
\bibitem{8363930}
\bibinfo{author}{Z.~\surnamestart {Chen}\surnameend}, \bibinfo{author}{C.~K.
  \surnamestart {Yeo}\surnameend}, \bibinfo{author}{B.~S. \surnamestart
  {Lee}\surnameend} \& \bibinfo{author}{C.~T. \surnamestart {Lau}\surnameend}
  (\bibinfo{year}{2018}): \emph{\bibinfo{title}{Autoencoder-based network
  anomaly detection}}.
\newblock In: {\sl \bibinfo{booktitle}{2018 Wireless Telecommunications
  Symposium (WTS)}}, pp. \bibinfo{pages}{1--5}, \doi{10.1109/WTS.2018.8363930}.

\bibitemdeclare{article}{dutta2018learning}
\bibitem{dutta2018learning}
\bibinfo{author}{Souradeep \surnamestart Dutta\surnameend},
  \bibinfo{author}{Susmit \surnamestart Jha\surnameend},
  \bibinfo{author}{Sriram \surnamestart Sankaranarayanan\surnameend} \&
  \bibinfo{author}{Ashish \surnamestart Tiwari\surnameend}
  (\bibinfo{year}{2018}): \emph{\bibinfo{title}{Learning and verification of
  feedback control systems using feedforward neural networks}}.
\newblock {\sl \bibinfo{journal}{IFAC-PapersOnLine}}
  \bibinfo{volume}{51}(\bibinfo{number}{16}), pp. \bibinfo{pages}{151--156},
  \doi{10.1016/j.ifacol.2018.08.026}.

\bibitemdeclare{inproceedings}{dutta2018output}
\bibitem{dutta2018output}
\bibinfo{author}{Souradeep \surnamestart Dutta\surnameend},
  \bibinfo{author}{Susmit \surnamestart Jha\surnameend},
  \bibinfo{author}{Sriram \surnamestart Sankaranarayanan\surnameend} \&
  \bibinfo{author}{Ashish \surnamestart Tiwari\surnameend}
  (\bibinfo{year}{2018}): \emph{\bibinfo{title}{Output range analysis for deep
  feedforward neural networks}}.
\newblock In: {\sl \bibinfo{booktitle}{NASA Formal Methods Symposium}},
  \bibinfo{organization}{Springer}, pp. \bibinfo{pages}{121--138},
  \doi{10.1007/978-3-319-77935-5\_9}.

\bibitemdeclare{inproceedings}{dvijotham2020efficient}
\bibitem{dvijotham2020efficient}
\bibinfo{author}{Krishnamurthy~Dj \surnamestart Dvijotham\surnameend},
  \bibinfo{author}{Robert \surnamestart Stanforth\surnameend},
  \bibinfo{author}{Sven \surnamestart Gowal\surnameend},
  \bibinfo{author}{Chongli \surnamestart Qin\surnameend},
  \bibinfo{author}{Soham \surnamestart De\surnameend} \&
  \bibinfo{author}{Pushmeet \surnamestart Kohli\surnameend}
  (\bibinfo{year}{2020}): \emph{\bibinfo{title}{Efficient neural network
  verification with exactness characterization}}.
\newblock In: {\sl \bibinfo{booktitle}{Uncertainty in Artificial
  Intelligence}}, \bibinfo{organization}{PMLR}, pp. \bibinfo{pages}{497--507}.

\bibitemdeclare{inproceedings}{ehlers2017formal}
\bibitem{ehlers2017formal}
\bibinfo{author}{Ruediger \surnamestart Ehlers\surnameend}
  (\bibinfo{year}{2017}): \emph{\bibinfo{title}{Formal verification of
  piece-wise linear feed-forward neural networks}}.
\newblock In: {\sl \bibinfo{booktitle}{International Symposium on Automated
  Technology for Verification and Analysis}}, \bibinfo{organization}{Springer},
  pp. \bibinfo{pages}{269--286}, \doi{10.1007/978-3-319-68167-2\_19}.

\bibitemdeclare{article}{elattar2019conception}
\bibitem{elattar2019conception}
\bibinfo{author}{Hatem~M \surnamestart Elattar\surnameend},
  \bibinfo{author}{Hamdy~K \surnamestart Elminir\surnameend} \&
  \bibinfo{author}{Alaa~Mohamed \surnamestart Riad\surnameend}
  (\bibinfo{year}{2019}): \emph{\bibinfo{title}{Conception and implementation
  of a data-driven prognostics algorithm for safety--critical systems}}.
\newblock {\sl \bibinfo{journal}{Soft Computing}}
  \bibinfo{volume}{23}(\bibinfo{number}{10}), pp. \bibinfo{pages}{3365--3382},
  \doi{10.1007/s00500-017-2995-7}.

\bibitemdeclare{misc}{full2020studying}
\bibitem{full2020studying}
\bibinfo{author}{Peter~M. \surnamestart Full\surnameend},
  \bibinfo{author}{Fabian \surnamestart Isensee\surnameend},
  \bibinfo{author}{Paul~F. \surnamestart J\IeC{\"a}ger\surnameend} \&
  \bibinfo{author}{Klaus \surnamestart Maier-Hein\surnameend}
  (\bibinfo{year}{2020}): \emph{\bibinfo{title}{Studying Robustness of Semantic
  Segmentation under Domain Shift in cardiac MRI}},
  \doi{10.1007/978-3-030-68107-4\_24}.
\newblock \eprint{2011.07592}.

\bibitemdeclare{inproceedings}{gehr2018ai}
\bibitem{gehr2018ai}
\bibinfo{author}{Timon \surnamestart Gehr\surnameend}, \bibinfo{author}{Matthew
  \surnamestart Mirman\surnameend}, \bibinfo{author}{Dana \surnamestart
  Drachsler-Cohen\surnameend}, \bibinfo{author}{Petar \surnamestart
  Tsankov\surnameend}, \bibinfo{author}{Swarat \surnamestart
  Chaudhuri\surnameend} \& \bibinfo{author}{Martin \surnamestart
  Vechev\surnameend} (\bibinfo{year}{2018}): \emph{\bibinfo{title}{Ai 2: Safety
  and robustness certification of neural networks with abstract
  interpretation}}.
\newblock In: {\sl \bibinfo{booktitle}{Security and Privacy (SP), 2018 IEEE
  Symposium on}}, \doi{10.1109/SP.2018.00058}.

\bibitemdeclare{article}{goodfellow2014explaining}
\bibitem{goodfellow2014explaining}
\bibinfo{author}{Ian~J \surnamestart Goodfellow\surnameend},
  \bibinfo{author}{Jonathon \surnamestart Shlens\surnameend} \&
  \bibinfo{author}{Christian \surnamestart Szegedy\surnameend}
  (\bibinfo{year}{2014}): \emph{\bibinfo{title}{Explaining and harnessing
  adversarial examples}}.
\newblock {\sl \bibinfo{journal}{arXiv preprint arXiv:1412.6572}},
  \doi{10.48550/arXiv.1412.6572}.

\bibitemdeclare{inproceedings}{huang2017safety}
\bibitem{huang2017safety}
\bibinfo{author}{Xiaowei \surnamestart Huang\surnameend},
  \bibinfo{author}{Marta \surnamestart Kwiatkowska\surnameend},
  \bibinfo{author}{Sen \surnamestart Wang\surnameend} \& \bibinfo{author}{Min
  \surnamestart Wu\surnameend} (\bibinfo{year}{2017}):
  \emph{\bibinfo{title}{Safety verification of deep neural networks}}.
\newblock In: {\sl \bibinfo{booktitle}{International Conference on Computer
  Aided Verification}}, \bibinfo{organization}{Springer}, pp.
  \bibinfo{pages}{3--29}, \doi{10.1007/978-3-319-63387-9\_1}.

\bibitemdeclare{inproceedings}{katz2017reluplex}
\bibitem{katz2017reluplex}
\bibinfo{author}{Guy \surnamestart Katz\surnameend}, \bibinfo{author}{Clark
  \surnamestart Barrett\surnameend}, \bibinfo{author}{David~L \surnamestart
  Dill\surnameend}, \bibinfo{author}{Kyle \surnamestart Julian\surnameend} \&
  \bibinfo{author}{Mykel~J \surnamestart Kochenderfer\surnameend}
  (\bibinfo{year}{2017}): \emph{\bibinfo{title}{Reluplex: An efficient SMT
  solver for verifying deep neural networks}}.
\newblock In: {\sl \bibinfo{booktitle}{International Conference on Computer
  Aided Verification}}, \bibinfo{organization}{Springer}, pp.
  \bibinfo{pages}{97--117}, \doi{10.1007/978-3-319-63387-9\_5}.

\bibitemdeclare{inproceedings}{katz2019marabou}
\bibitem{katz2019marabou}
\bibinfo{author}{Guy \surnamestart Katz\surnameend}, \bibinfo{author}{Derek~A
  \surnamestart Huang\surnameend}, \bibinfo{author}{Duligur \surnamestart
  Ibeling\surnameend}, \bibinfo{author}{Kyle \surnamestart Julian\surnameend},
  \bibinfo{author}{Christopher \surnamestart Lazarus\surnameend},
  \bibinfo{author}{Rachel \surnamestart Lim\surnameend}, \bibinfo{author}{Parth
  \surnamestart Shah\surnameend}, \bibinfo{author}{Shantanu \surnamestart
  Thakoor\surnameend}, \bibinfo{author}{Haoze \surnamestart Wu\surnameend} \&
  \bibinfo{author}{Aleksandar \surnamestart Zelji{\'c}\surnameend}
  (\bibinfo{year}{2019}): \emph{\bibinfo{title}{The marabou framework for
  verification and analysis of deep neural networks}}.
\newblock In: {\sl \bibinfo{booktitle}{International Conference on Computer
  Aided Verification}}, \bibinfo{organization}{Springer}, pp.
  \bibinfo{pages}{443--452}, \doi{10.1007/978-3-030-25540-4\_26}.

\bibitemdeclare{article}{khmelnitsky2020property}
\bibitem{khmelnitsky2020property}
\bibinfo{author}{Igor \surnamestart Khmelnitsky\surnameend},
  \bibinfo{author}{Daniel \surnamestart Neider\surnameend},
  \bibinfo{author}{Rajarshi \surnamestart Roy\surnameend},
  \bibinfo{author}{Beno{\^\i}t \surnamestart Barbot\surnameend},
  \bibinfo{author}{Benedikt \surnamestart Bollig\surnameend},
  \bibinfo{author}{Alain \surnamestart Finkel\surnameend},
  \bibinfo{author}{Serge \surnamestart Haddad\surnameend},
  \bibinfo{author}{Martin \surnamestart Leucker\surnameend} \&
  \bibinfo{author}{Lina \surnamestart Ye\surnameend} (\bibinfo{year}{2020}):
  \emph{\bibinfo{title}{Property-Directed Verification of Recurrent Neural
  Networks}}.
\newblock {\sl \bibinfo{journal}{arXiv preprint arXiv:2009.10610}},
  \doi{10.48550/arXiv.2009.10610}.

\bibitemdeclare{inproceedings}{Klingner}
\bibitem{Klingner}
\bibinfo{author}{Marvin \surnamestart Klingner\surnameend},
  \bibinfo{author}{Andreas \surnamestart Bar\surnameend} \&
  \bibinfo{author}{Tim \surnamestart Fingscheidt\surnameend}
  (\bibinfo{year}{2020}): \emph{\bibinfo{title}{Improved Noise and Attack
  Robustness for Semantic Segmentation by Using Multi-Task Training With
  Self-Supervised Depth Estimation}}.
\newblock In: {\sl \bibinfo{booktitle}{Proceedings of the IEEE/CVF Conference
  on Computer Vision and Pattern Recognition (CVPR) Workshops}},
  \doi{10.1109/CVPRW50498.2020.00168}.

\bibitemdeclare{article}{kouvaros2018formal}
\bibitem{kouvaros2018formal}
\bibinfo{author}{Panagiotis \surnamestart Kouvaros\surnameend} \&
  \bibinfo{author}{Alessio \surnamestart Lomuscio\surnameend}
  (\bibinfo{year}{2018}): \emph{\bibinfo{title}{Formal verification of
  cnn-based perception systems}}.
\newblock {\sl \bibinfo{journal}{arXiv preprint arXiv:1811.11373}},
  \doi{10.48550/arXiv.1811.11373}.

\bibitemdeclare{article}{lomuscio2017approach}
\bibitem{lomuscio2017approach}
\bibinfo{author}{Alessio \surnamestart Lomuscio\surnameend} \&
  \bibinfo{author}{Lalit \surnamestart Maganti\surnameend}
  (\bibinfo{year}{2017}): \emph{\bibinfo{title}{An approach to reachability
  analysis for feed-forward relu neural networks}}.
\newblock {\sl \bibinfo{journal}{arXiv preprint arXiv:1706.07351}},
  \doi{10.48550/arXiv.1706.07351}.

\bibitemdeclare{misc}{2020Segmentation}
\bibitem{2020Segmentation}
\bibinfo{author}{Shervin \surnamestart Minaee\surnameend},
  \bibinfo{author}{Yuri \surnamestart Boykov\surnameend},
  \bibinfo{author}{Fatih \surnamestart Porikli\surnameend},
  \bibinfo{author}{Antonio \surnamestart Plaza\surnameend},
  \bibinfo{author}{Nasser \surnamestart Kehtarnavaz\surnameend} \&
  \bibinfo{author}{Demetri \surnamestart Terzopoulos\surnameend}
  (\bibinfo{year}{2020}): \emph{\bibinfo{title}{Image Segmentation Using Deep
  Learning: A Survey}}, \doi{10.48550/arXiv.2001.05566}.
\newblock \eprint{2001.05566}.

\bibitemdeclare{inproceedings}{moosavi2016deepfool}
\bibitem{moosavi2016deepfool}
\bibinfo{author}{Seyed-Mohsen \surnamestart Moosavi-Dezfooli\surnameend},
  \bibinfo{author}{Alhussein \surnamestart Fawzi\surnameend} \&
  \bibinfo{author}{Pascal \surnamestart Frossard\surnameend}
  (\bibinfo{year}{2016}): \emph{\bibinfo{title}{Deepfool: a simple and accurate
  method to fool deep neural networks}}.
\newblock In: {\sl \bibinfo{booktitle}{Proceedings of the IEEE conference on
  computer vision and pattern recognition}}, pp. \bibinfo{pages}{2574--2582},
  \doi{10.1109/CVPR.2016.282}.

\bibitemdeclare{article}{munir2020performance}
\bibitem{munir2020performance}
\bibinfo{author}{Nauman \surnamestart Munir\surnameend},
  \bibinfo{author}{Jinhyun \surnamestart Park\surnameend},
  \bibinfo{author}{Hak-Joon \surnamestart Kim\surnameend},
  \bibinfo{author}{Sung-Jin \surnamestart Song\surnameend} \&
  \bibinfo{author}{Sung-Sik \surnamestart Kang\surnameend}
  (\bibinfo{year}{2020}): \emph{\bibinfo{title}{Performance enhancement of
  convolutional neural network for ultrasonic flaw classification by adopting
  autoencoder}}.
\newblock {\sl \bibinfo{journal}{NDT \& E International}}
  \bibinfo{volume}{111}, p. \bibinfo{pages}{102218},
  \doi{10.1016/j.ndteint.2020.102218}.

\bibitemdeclare{article}{Oliveira2017}
\bibitem{Oliveira2017}
\bibinfo{author}{Gabriel \surnamestart Oliveira\surnameend},
  \bibinfo{author}{Claas \surnamestart Bollen\surnameend},
  \bibinfo{author}{Wolfram \surnamestart Burgard\surnameend} \&
  \bibinfo{author}{Thomas \surnamestart Brox\surnameend}
  (\bibinfo{year}{2017}): \emph{\bibinfo{title}{Efficient and robust deep
  networks for semantic segmentation}}.
\newblock {\sl \bibinfo{journal}{The International Journal of Robotics
  Research}} \bibinfo{volume}{37}, p. \bibinfo{pages}{027836491771054},
  \doi{10.1177/0278364917710542}.

\bibitemdeclare{inproceedings}{pulina2010abstraction}
\bibitem{pulina2010abstraction}
\bibinfo{author}{Luca \surnamestart Pulina\surnameend} \&
  \bibinfo{author}{Armando \surnamestart Tacchella\surnameend}
  (\bibinfo{year}{2010}): \emph{\bibinfo{title}{An abstraction-refinement
  approach to verification of artificial neural networks}}.
\newblock In: {\sl \bibinfo{booktitle}{International Conference on Computer
  Aided Verification}}, \bibinfo{organization}{Springer}, pp.
  \bibinfo{pages}{243--257}, \doi{10.1007/978-3-642-14295-6\_24}.

\bibitemdeclare{article}{ringner2008principal}
\bibitem{ringner2008principal}
\bibinfo{author}{Markus \surnamestart Ringn{\'e}r\surnameend}
  (\bibinfo{year}{2008}): \emph{\bibinfo{title}{What is principal component
  analysis?}}
\newblock {\sl \bibinfo{journal}{Nature biotechnology}}
  \bibinfo{volume}{26}(\bibinfo{number}{3}), pp. \bibinfo{pages}{303--304},
  \doi{10.1038/nbt0308-303}.

\bibitemdeclare{inproceedings}{sakurada2014anomaly}
\bibitem{sakurada2014anomaly}
\bibinfo{author}{Mayu \surnamestart Sakurada\surnameend} \&
  \bibinfo{author}{Takehisa \surnamestart Yairi\surnameend}
  (\bibinfo{year}{2014}): \emph{\bibinfo{title}{Anomaly detection using
  autoencoders with nonlinear dimensionality reduction}}.
\newblock In: {\sl \bibinfo{booktitle}{Proceedings of the MLSDA 2014 2nd
  Workshop on Machine Learning for Sensory Data Analysis}}, pp.
  \bibinfo{pages}{4--11}, \doi{10.1145/2689746.2689747}.

\bibitemdeclare{article}{saxena2017design}
\bibitem{saxena2017design}
\bibinfo{author}{Divya \surnamestart Saxena\surnameend} \&
  \bibinfo{author}{Vaskar \surnamestart Raychoudhury\surnameend}
  (\bibinfo{year}{2017}): \emph{\bibinfo{title}{Design and verification of an
  NDN-based safety-critical application: A case study with smart healthcare}}.
\newblock {\sl \bibinfo{journal}{ieee transactions on systems, man, and
  cybernetics: systems}} \bibinfo{volume}{49}(\bibinfo{number}{5}), pp.
  \bibinfo{pages}{991--1005}, \doi{10.1109/TSMC.2017.2723843}.

\bibitemdeclare{inproceedings}{singh2018fast}
\bibitem{singh2018fast}
\bibinfo{author}{Gagandeep \surnamestart Singh\surnameend},
  \bibinfo{author}{Timon \surnamestart Gehr\surnameend},
  \bibinfo{author}{Matthew \surnamestart Mirman\surnameend},
  \bibinfo{author}{Markus \surnamestart P{\"u}schel\surnameend} \&
  \bibinfo{author}{Martin \surnamestart Vechev\surnameend}
  (\bibinfo{year}{2018}): \emph{\bibinfo{title}{Fast and effective robustness
  certification}}.
\newblock In: {\sl \bibinfo{booktitle}{Advances in Neural Information
  Processing Systems}}, pp. \bibinfo{pages}{10825--10836}.

\bibitemdeclare{article}{singh2019abstract}
\bibitem{singh2019abstract}
\bibinfo{author}{Gagandeep \surnamestart Singh\surnameend},
  \bibinfo{author}{Timon \surnamestart Gehr\surnameend},
  \bibinfo{author}{Markus \surnamestart P{\"u}schel\surnameend} \&
  \bibinfo{author}{Martin \surnamestart Vechev\surnameend}
  (\bibinfo{year}{2019}): \emph{\bibinfo{title}{An abstract domain for
  certifying neural networks}}.
\newblock {\sl \bibinfo{journal}{Proceedings of the ACM on Programming
  Languages}} \bibinfo{volume}{3}(\bibinfo{number}{POPL}),
  p.~\bibinfo{pages}{41}, \doi{10.1145/3290354}.

\bibitemdeclare{article}{sun2016sparse}
\bibitem{sun2016sparse}
\bibinfo{author}{Wenjun \surnamestart Sun\surnameend}, \bibinfo{author}{Siyu
  \surnamestart Shao\surnameend}, \bibinfo{author}{Rui \surnamestart
  Zhao\surnameend}, \bibinfo{author}{Ruqiang \surnamestart Yan\surnameend},
  \bibinfo{author}{Xingwu \surnamestart Zhang\surnameend} \&
  \bibinfo{author}{Xuefeng \surnamestart Chen\surnameend}
  (\bibinfo{year}{2016}): \emph{\bibinfo{title}{A sparse auto-encoder-based
  deep neural network approach for induction motor faults classification}}.
\newblock {\sl \bibinfo{journal}{Measurement}} \bibinfo{volume}{89}, pp.
  \bibinfo{pages}{171--178}, \doi{10.1016/j.measurement.2016.04.007}.

\bibitemdeclare{article}{LBFGS}
\bibitem{LBFGS}
\bibinfo{author}{Christian \surnamestart Szegedy\surnameend},
  \bibinfo{author}{Wojciech \surnamestart Zaremba\surnameend},
  \bibinfo{author}{Ilya \surnamestart Sutskever\surnameend},
  \bibinfo{author}{Joan \surnamestart Bruna\surnameend},
  \bibinfo{author}{Dumitru \surnamestart Erhan\surnameend},
  \bibinfo{author}{Ian \surnamestart Goodfellow\surnameend} \&
  \bibinfo{author}{Rob \surnamestart Fergus\surnameend} (\bibinfo{year}{2013}):
  \emph{\bibinfo{title}{Intriguing properties of neural networks}}.
\newblock {\sl \bibinfo{journal}{arXiv preprint arXiv:1312.6199}},
  \doi{10.48550/arXiv.1312.6199}.

\bibitemdeclare{inproceedings}{tran2020verification}
\bibitem{tran2020verification}
\bibinfo{author}{Hoang-Dung \surnamestart Tran\surnameend},
  \bibinfo{author}{Stanley \surnamestart Bak\surnameend},
  \bibinfo{author}{Weiming \surnamestart Xiang\surnameend} \&
  \bibinfo{author}{Taylor~T \surnamestart Johnson\surnameend}
  (\bibinfo{year}{2020}): \emph{\bibinfo{title}{Verification of deep
  convolutional neural networks using imagestars}}.
\newblock In: {\sl \bibinfo{booktitle}{International Conference on Computer
  Aided Verification}}, \bibinfo{organization}{Springer}, pp.
  \bibinfo{pages}{18--42}, \doi{10.1007/978-3-030-53288-8\_2}.

\bibitemdeclare{inproceedings}{tran2019parallel}
\bibitem{tran2019parallel}
\bibinfo{author}{Hoang-Dung \surnamestart Tran\surnameend},
  \bibinfo{author}{Patrick \surnamestart Musau\surnameend},
  \bibinfo{author}{Diego~Manzanas \surnamestart Lopez\surnameend},
  \bibinfo{author}{Xiaodong \surnamestart Yang\surnameend},
  \bibinfo{author}{Luan~Viet \surnamestart Nguyen\surnameend},
  \bibinfo{author}{Weiming \surnamestart Xiang\surnameend} \&
  \bibinfo{author}{Taylor~T. \surnamestart Johnson\surnameend}
  (\bibinfo{year}{2019}): \emph{\bibinfo{title}{Parallelizable Reachability
  Analysis Algorithms for Feed-Forward Neural Networks}}.
\newblock In: {\sl \bibinfo{booktitle}{7th International Conference on Formal
  Methods in Software Engineering (FormaliSE2019), Montreal, Canada}},
  \doi{10.1109/FormaliSE.2019.00012}.

\bibitemdeclare{inproceedings}{tran2019fm}
\bibitem{tran2019fm}
\bibinfo{author}{Hoang-Dung \surnamestart Tran\surnameend},
  \bibinfo{author}{Patrick \surnamestart Musau\surnameend},
  \bibinfo{author}{Diego~Manzanas \surnamestart Lopez\surnameend},
  \bibinfo{author}{Xiaodong \surnamestart Yang\surnameend},
  \bibinfo{author}{Luan~Viet \surnamestart Nguyen\surnameend},
  \bibinfo{author}{Weiming \surnamestart Xiang\surnameend} \&
  \bibinfo{author}{Taylor~T. \surnamestart Johnson\surnameend}
  (\bibinfo{year}{2019}): \emph{\bibinfo{title}{Star-Based Reachability
  Analysis for Deep Neural Networks}}.
\newblock In: {\sl \bibinfo{booktitle}{23rd International Symposisum on Formal
  Methods (FM'19)}}, \bibinfo{publisher}{Springer International Publishing},
  \doi{10.1007/978-3-030-30942-8_39}.

\bibitemdeclare{inproceedings}{tran2021cav}
\bibitem{tran2021cav}
\bibinfo{author}{Hoang-Dung \surnamestart Tran\surnameend},
  \bibinfo{author}{Neelanjana \surnamestart Pal\surnameend},
  \bibinfo{author}{Patrick \surnamestart Musau\surnameend},
  \bibinfo{author}{Xiaodong \surnamestart Yang\surnameend},
  \bibinfo{author}{Nathaniel~P \surnamestart Hamilton\surnameend},
  \bibinfo{author}{Diego~Manzanas \surnamestart Lopez\surnameend},
  \bibinfo{author}{Stanley \surnamestart Bak\surnameend} \&
  \bibinfo{author}{Taylor~T \surnamestart Johnson\surnameend}
  (\bibinfo{year}{2021}): \emph{\bibinfo{title}{Robustness Verification of
  Semantic Segmentation Neural Networks using Relaxed Reachability}}.
\newblock In: {\sl \bibinfo{booktitle}{Proceedings of the 33rd International
  Conference on Computer-Aided Verification}}, \bibinfo{publisher}{Springer},
  \doi{10.1007/978-3-030-81685-8\_12}.

\bibitemdeclare{inproceedings}{tran2020nnv}
\bibitem{tran2020nnv}
\bibinfo{author}{Hoang-Dung \surnamestart Tran\surnameend},
  \bibinfo{author}{Xiaodong \surnamestart Yang\surnameend},
  \bibinfo{author}{Diego~Manzanas \surnamestart Lopez\surnameend},
  \bibinfo{author}{Patrick \surnamestart Musau\surnameend},
  \bibinfo{author}{Luan~Viet \surnamestart Nguyen\surnameend},
  \bibinfo{author}{Weiming \surnamestart Xiang\surnameend},
  \bibinfo{author}{Stanley \surnamestart Bak\surnameend} \&
  \bibinfo{author}{Taylor~T \surnamestart Johnson\surnameend}
  (\bibinfo{year}{2020}): \emph{\bibinfo{title}{NNV: The neural network
  verification tool for deep neural networks and learning-enabled
  cyber-physical systems}}.
\newblock In: {\sl \bibinfo{booktitle}{International Conference on Computer
  Aided Verification}}, \bibinfo{organization}{Springer}, pp.
  \bibinfo{pages}{3--17}, \doi{10.1007/978-3-030-53288-8\_1}.

\bibitemdeclare{inproceedings}{vincent2008extracting}
\bibitem{vincent2008extracting}
\bibinfo{author}{Pascal \surnamestart Vincent\surnameend},
  \bibinfo{author}{Hugo \surnamestart Larochelle\surnameend},
  \bibinfo{author}{Yoshua \surnamestart Bengio\surnameend} \&
  \bibinfo{author}{Pierre-Antoine \surnamestart Manzagol\surnameend}
  (\bibinfo{year}{2008}): \emph{\bibinfo{title}{Extracting and composing robust
  features with denoising autoencoders}}.
\newblock In: {\sl \bibinfo{booktitle}{Proceedings of the 25th international
  conference on Machine learning}}, pp. \bibinfo{pages}{1096--1103},
  \doi{10.1145/1390156.1390294}.

\bibitemdeclare{article}{vincent2010stacked}
\bibitem{vincent2010stacked}
\bibinfo{author}{Pascal \surnamestart Vincent\surnameend},
  \bibinfo{author}{Hugo \surnamestart Larochelle\surnameend},
  \bibinfo{author}{Isabelle \surnamestart Lajoie\surnameend},
  \bibinfo{author}{Yoshua \surnamestart Bengio\surnameend},
  \bibinfo{author}{Pierre-Antoine \surnamestart Manzagol\surnameend} \&
  \bibinfo{author}{L{\'e}on \surnamestart Bottou\surnameend}
  (\bibinfo{year}{2010}): \emph{\bibinfo{title}{Stacked denoising autoencoders:
  Learning useful representations in a deep network with a local denoising
  criterion.}}
\newblock {\sl \bibinfo{journal}{Journal of machine learning research}}
  \bibinfo{volume}{11}(\bibinfo{number}{12}).

\bibitemdeclare{inproceedings}{wang2018efficient}
\bibitem{wang2018efficient}
\bibinfo{author}{Shiqi \surnamestart Wang\surnameend}, \bibinfo{author}{Kexin
  \surnamestart Pei\surnameend}, \bibinfo{author}{Justin \surnamestart
  Whitehouse\surnameend}, \bibinfo{author}{Junfeng \surnamestart
  Yang\surnameend} \& \bibinfo{author}{Suman \surnamestart Jana\surnameend}
  (\bibinfo{year}{2018}): \emph{\bibinfo{title}{Efficient formal safety
  analysis of neural networks}}.
\newblock In: {\sl \bibinfo{booktitle}{Advances in Neural Information
  Processing Systems}}, pp. \bibinfo{pages}{6369--6379},
  \doi{10.48550/arXiv.1809.08098}.

\bibitemdeclare{article}{wang2018formal}
\bibitem{wang2018formal}
\bibinfo{author}{Shiqi \surnamestart Wang\surnameend}, \bibinfo{author}{Kexin
  \surnamestart Pei\surnameend}, \bibinfo{author}{Justin \surnamestart
  Whitehouse\surnameend}, \bibinfo{author}{Junfeng \surnamestart
  Yang\surnameend} \& \bibinfo{author}{Suman \surnamestart Jana\surnameend}
  (\bibinfo{year}{2018}): \emph{\bibinfo{title}{Formal Security Analysis of
  Neural Networks using Symbolic Intervals}}.
\newblock {\sl \bibinfo{journal}{arXiv preprint arXiv:1804.10829}},
  \doi{10.48550/arXiv.1804.10829}.

\bibitemdeclare{article}{weng2018towards}
\bibitem{weng2018towards}
\bibinfo{author}{Tsui-Wei \surnamestart Weng\surnameend}, \bibinfo{author}{Huan
  \surnamestart Zhang\surnameend}, \bibinfo{author}{Hongge \surnamestart
  Chen\surnameend}, \bibinfo{author}{Zhao \surnamestart Song\surnameend},
  \bibinfo{author}{Cho-Jui \surnamestart Hsieh\surnameend},
  \bibinfo{author}{Duane \surnamestart Boning\surnameend},
  \bibinfo{author}{Inderjit~S \surnamestart Dhillon\surnameend} \&
  \bibinfo{author}{Luca \surnamestart Daniel\surnameend}
  (\bibinfo{year}{2018}): \emph{\bibinfo{title}{Towards Fast Computation of
  Certified Robustness for ReLU Networks}}.
\newblock {\sl \bibinfo{journal}{arXiv preprint arXiv:1804.09699}},
  \doi{10.48550/arXiv.1804.09699}.

\bibitemdeclare{article}{xiang2017reachable}
\bibitem{xiang2017reachable}
\bibinfo{author}{Weiming \surnamestart Xiang\surnameend},
  \bibinfo{author}{Hoang-Dung \surnamestart Tran\surnameend} \&
  \bibinfo{author}{Taylor~T \surnamestart Johnson\surnameend}
  (\bibinfo{year}{2017}): \emph{\bibinfo{title}{Reachable set computation and
  safety verification for neural networks with ReLU activations}}.
\newblock {\sl \bibinfo{journal}{arXiv preprint arXiv:1712.08163}},
  \doi{10.48550/arXiv.1712.08163}.

\bibitemdeclare{article}{xiang2018output}
\bibitem{xiang2018output}
\bibinfo{author}{Weiming \surnamestart Xiang\surnameend},
  \bibinfo{author}{Hoang-Dung \surnamestart Tran\surnameend} \&
  \bibinfo{author}{Taylor~T \surnamestart Johnson\surnameend}
  (\bibinfo{year}{2018}): \emph{\bibinfo{title}{Output reachable set estimation
  and verification for multilayer neural networks}}.
\newblock {\sl \bibinfo{journal}{IEEE transactions on neural networks and
  learning systems}} \bibinfo{volume}{29}(\bibinfo{number}{11}), pp.
  \bibinfo{pages}{5777--5783}, \doi{10.1109/TNNLS.2018.2808470}.

\bibitemdeclare{article}{xiang2018specification}
\bibitem{xiang2018specification}
\bibinfo{author}{Weiming \surnamestart Xiang\surnameend},
  \bibinfo{author}{Hoang-Dung \surnamestart Tran\surnameend} \&
  \bibinfo{author}{Taylor~T \surnamestart Johnson\surnameend}
  (\bibinfo{year}{2019}): \emph{\bibinfo{title}{Specification-Guided Safety
  Verification for Feedforward Neural Networks}}.
\newblock {\sl \bibinfo{journal}{AAAI Spring Symposium on Verification of
  Neural Networks}}, \doi{10.48550/arXiv.1812.06161}.

\bibitemdeclare{inproceedings}{7449810}
\bibitem{7449810}
\bibinfo{author}{C.~\surnamestart {Zhang}\surnameend},
  \bibinfo{author}{W.~\surnamestart {Gao}\surnameend},
  \bibinfo{author}{J.~\surnamestart {Song}\surnameend} \&
  \bibinfo{author}{J.~\surnamestart {Jiang}\surnameend} (\bibinfo{year}{2016}):
  \emph{\bibinfo{title}{An imbalanced data classification algorithm of improved
  autoencoder neural network}}.
\newblock In: {\sl \bibinfo{booktitle}{2016 Eighth International Conference on
  Advanced Computational Intelligence (ICACI)}}, pp. \bibinfo{pages}{95--99},
  \doi{10.1109/ICACI.2016.7449810}.

\bibitemdeclare{inproceedings}{zhang2018efficient}
\bibitem{zhang2018efficient}
\bibinfo{author}{Huan \surnamestart Zhang\surnameend},
  \bibinfo{author}{Tsui-Wei \surnamestart Weng\surnameend},
  \bibinfo{author}{Pin-Yu \surnamestart Chen\surnameend},
  \bibinfo{author}{Cho-Jui \surnamestart Hsieh\surnameend} \&
  \bibinfo{author}{Luca \surnamestart Daniel\surnameend}
  (\bibinfo{year}{2018}): \emph{\bibinfo{title}{Efficient neural network
  robustness certification with general activation functions}}.
\newblock In: {\sl \bibinfo{booktitle}{Advances in Neural Information
  Processing Systems}}, pp. \bibinfo{pages}{4944--4953},
  \doi{10.48550/arXiv.1811.00866}.

\bibitemdeclare{inproceedings}{zhou2017anomaly}
\bibitem{zhou2017anomaly}
\bibinfo{author}{Chong \surnamestart Zhou\surnameend} \&
  \bibinfo{author}{Randy~C \surnamestart Paffenroth\surnameend}
  (\bibinfo{year}{2017}): \emph{\bibinfo{title}{Anomaly detection with robust
  deep autoencoders}}.
\newblock In: {\sl \bibinfo{booktitle}{Proceedings of the 23rd ACM SIGKDD
  international conference on knowledge discovery and data mining}}, pp.
  \bibinfo{pages}{665--674}, \doi{10.1145/3097983.3098052}.

\bibitemdeclare{article}{Zhou2020AutomatedEO}
\bibitem{Zhou2020AutomatedEO}
\bibinfo{author}{W.~\surnamestart Zhou\surnameend},
  \bibinfo{author}{J.~\surnamestart Berrio\surnameend},
  \bibinfo{author}{S.~\surnamestart Worrall\surnameend} \&
  \bibinfo{author}{Eduardo~M. \surnamestart Nebot\surnameend}
  (\bibinfo{year}{2020}): \emph{\bibinfo{title}{Automated Evaluation of
  Semantic Segmentation Robustness for Autonomous Driving}}.
\newblock {\sl \bibinfo{journal}{IEEE Transactions on Intelligent
  Transportation Systems}} \bibinfo{volume}{21}, pp.
  \bibinfo{pages}{1951--1963}, \doi{10.1109/TITS.2019.2909066}.

\end{thebibliography}
